\documentclass{article}

\usepackage{microtype}
\usepackage{graphicx}
\usepackage{subcaption}
\usepackage{booktabs}
\usepackage{orcidlink}
\usepackage{hyperref}

\usepackage[preprint]{arxiv2026}

\usepackage{amsmath}
\usepackage{amssymb}
\usepackage{mathtools}
\usepackage{amsthm}

\usepackage[capitalize,noabbrev]{cleveref}

\theoremstyle{plain}

\theoremstyle{definition}

\theoremstyle{remark}

\usepackage[textsize=tiny]{todonotes}

\arxivtitlerunning{Accelerating Pulsar Posterior Inference via Learned Latent Representations and Local Optimization}

\begin{document}

\twocolumn
  [\arxivtitle{Accelerating Posterior Inference from Pulsar Light Curves via Learned \\ Latent Representations and Local Simulator-Guided Optimization}

  \arxivsetsymbol{equal}{*}

  \begin{arxivauthorlist}
    \arxivauthor{Farhana Taiyebah}{lab,sch} \orcidlink{0000-0003-3481-5913} \quad
    \arxivauthor{Abu Bucker Siddik}{lab} \orcidlink{0009-0009-4871-7231} \quad
    \arxivauthor{Indronil Bhattacharjee}{lab,nmsu} \orcidlink{0000-0002-3463-3389} \quad
    \arxivauthor{Diane Oyen}{lab} \orcidlink{0000-0002-1353-3688} \quad
    \\
    \arxivauthor{Soumi De}{lab} \orcidlink{0000-0002-3316-5149} \quad
    \arxivauthor{Greg Olmschenk}{nasa} \orcidlink{0000-0001-8472-2219} \quad
    \arxivauthor{Constantinos Kalapotharakos}{nasa} \orcidlink{0000-0003-1080-5286}
    
  \end{arxivauthorlist}
  
  \arxivaffiliation{lab}{Los Alamos National Laboratory, Los Alamos, NM, USA}
  \arxivaffiliation{nasa}{NASA Goddard Space Flight Center, Greenbelt, MD, USA}
  \arxivaffiliation{sch}{Department of Scientific Computing, Florida State University, Tallahassee, FL, USA}
  \arxivaffiliation{nmsu}{Department of Computer Science, New Mexico State University, Las Cruces, NM, USA}

  \vskip 0.05in
  \printAffiliations{}
  \vskip 0.15in
]

% \twocolumn
% \printAffiliationsAndNotice{}
\begin{abstractarxiv}
  Posterior inference from pulsar observations in the form of light curves is commonly performed using Markov chain Monte Carlo methods, which are accurate but computationally expensive. We introduce a framework that accelerates posterior inference while maintaining accuracy by combining learned latent representations with local simulator-guided optimization. A masked U-Net is first pretrained to reconstruct complete light curves from partial observations and to produce informative latent embeddings. Given a query light curve, we identify similar simulated light curves from the simulation bank by measuring similarity in the learned embedding space produced by pretrained U-Net encoder, yielding an initial empirical approximation to the posterior over parameters. This initialization is then refined using a local optimization procedure using hill-climbing updates, guided by a forward simulator, progressively shifting the empirical posterior toward higher-likelihood parameter regions. Experiments on the observed light curve of PSR J0030+0451, captured by NASA’s Neutron Star Interior Composition Explorer (NICER), show that our method closely matches posterior estimates obtained using traditional MCMC methods while achieving $\mathbf{120\times}$ reduction in inference time (from 24 hours to 12 minutes), demonstrating the effectiveness of learned representations and simulator-guided optimization for accelerated posterior inference.
\end{abstractarxiv}

\section{Introduction}
% problem
Pulsars are rapidly rotating neutron stars that emit beams of electromagnetic radiation, observed as periodic pulses when aligned with the Earth \cite{sturrock1971, smith1977}. Characterizing pulsar properties--such as mass, radius, surface temperature, and magnetic field geometry \cite{lattimer2016neutron,ozel2016masses}--requires solving an inverse problem: inferring the posterior distribution over physical parameters given observed light curves.
Traditional Markov chain Monte Carlo (MCMC) sampling \cite{metropolis1953equation, hastings1970monte, gilks1995markov} provides accurate posterior estimates \cite{constantinos2021} but is computationally very expensive, often requiring days to weeks on thousands of CPUs due to the high-dimensional parameter space and parameter degeneracies that create complex, multimodal posterior landscapes. \citet{greg2025} recently developed a convolutional neural network based simulator that replaces the expensive physical simulator within MCMC for pulsar posterior estimation, achieving a $>400\times$ speedup and enabling convergence in approximately one day on $4000$ CPUs--a task that would otherwise require over a year. However, even this accelerated approach still requires substantial computation time and remains constrained by the inherently iterative nature of MCMC sampling, which 
explores the parameter space through many local steps to adequately characterize strongly multimodal posteriors. \citet{siddik2025degeneracy} developed a Transformer–LSTM–based model
to estimate degenerate pulsar parameters from light curves with high posterior likelihood. However, this approach does not estimate the full posterior distribution and does not guarantee coverage of all high–posterior-likelihood regions in the parameter space.

%proposed solution
We introduce a four-stage inference framework that bypasses MCMC while preserving multimodal and degenerate posterior structure. Given a %query
observed light curve, the method retrieves other simulated light curves, that have similar characteristics, from a simulation bank using a learned embedding space. The parameters associated with the retrieved light curves define an empirical posterior approximation, which is subsequently refined via a simulator-guided hill-climbing procedure. This local optimization sharpens posterior mass within each region of support without collapsing distinct modes or erasing parameter degeneracies captured during the retrieval stage. While this method does not aim to generate exact posterior samples or provide asymptotic guarantees, it offers a computationally efficient approximation that preserves key posterior characteristics, such as multimodality and parameter degeneracies, and empirically aligns with MCMC results in settings where full Bayesian sampling is infeasible.

%contribution
The contributions of our work are the following: (\textbf{1}) a masked U-Net architecture that learns multi-scale latent representations of time-series data through self-supervised masked reconstruction, enabling similarity-based posterior approximation via retrieval in latent embedding space; (\textbf{2}) a hill-climbing optimization procedure that refines retrieved parameter sets through adaptive $P$-dimensional local search guided by the forward simulator, preserving multimodality while increasing posterior density; (\textbf{3}) experimental validation on PSR J0030+0451 demonstrating $120\times $ speedup over traditional MCMC with quantitative posterior agreement across all parameters; (\textbf{4}) demonstration of the proposed approach on a physics inference problem, specifically a gamma-ray spectrum (GRS) inference task \cite{moghani2025sampling}; and (\textbf{5}) ablation studies establishing that multi-level embeddings outperform both single-level embeddings and direct cosine similarity on raw light curves on retrieval task (Appendix \ref{app:ablation}).

\section{Methods}
\label{sec:methods}

Let $\mathcal{D}=\{(\mathbf{x}_i,\boldsymbol{\theta}_i)\}_{i=1}^N$ denote a simulation bank of $N$ light curves $\mathbf{x}_i\in\mathbb{R}^{T}$ and their parameters $\boldsymbol{\theta}_i\in\mathbb{R}^{P}$ (e.g. for pulsar parameter estimation, $T=64$ phase bins and $P=11$ parameters).

\textbf{Problem Statement.} Our goal is to infer the posterior distribution,  $p(\boldsymbol{\theta} | \mathbf{x}^*)$,  over physical parameters given an observed light curve $\mathbf{x}^*$. The proposed workflow for posterior prediction (see Figure~\ref{fig:method}) consists of:

\begin{figure*}[ht]
  \centering
  \includegraphics[width=\linewidth]{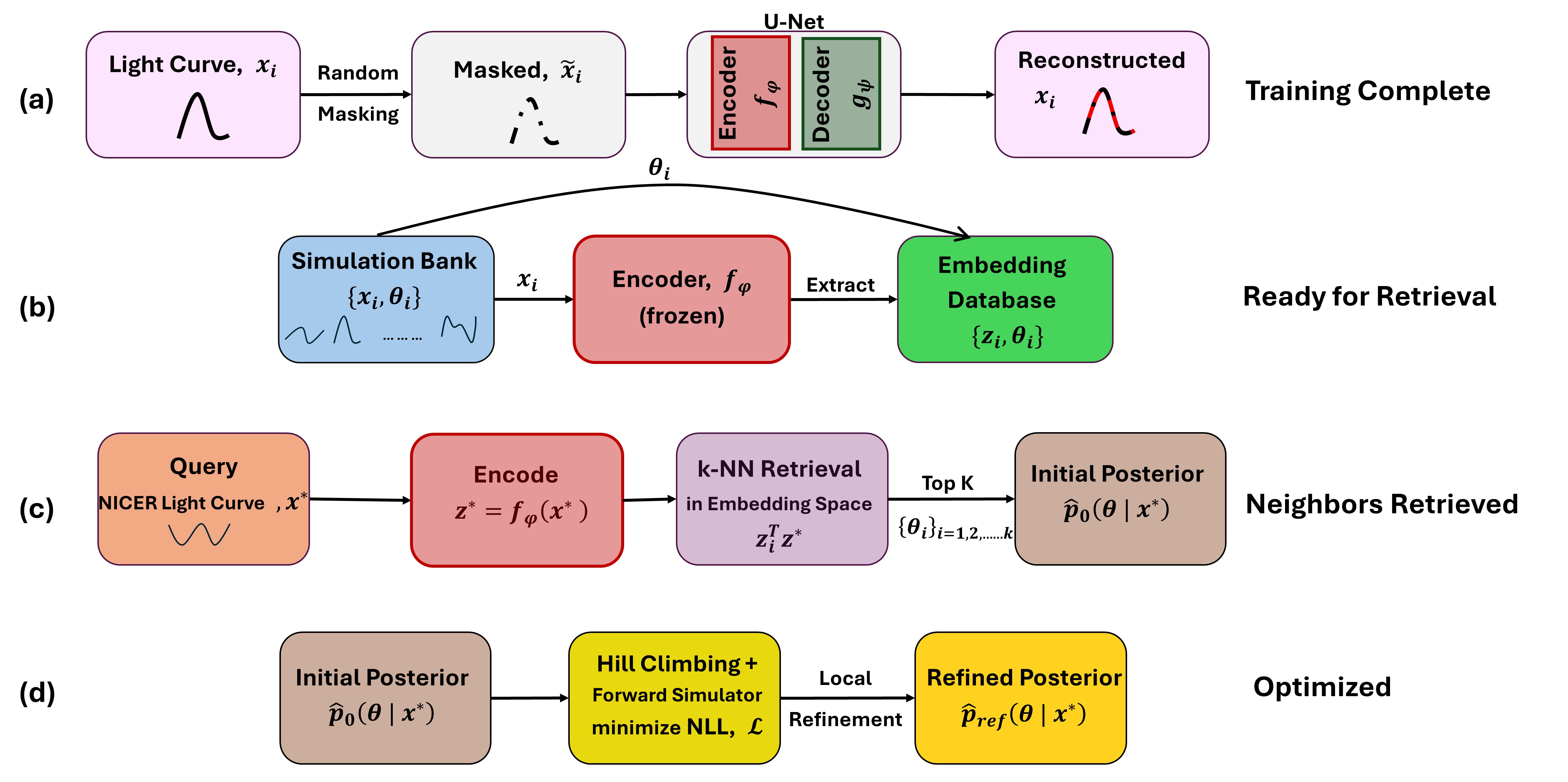}
  \caption{Method overview for accelerated posterior inference from pulsar light curves. \textbf{(a) Masked U-Net pretraining:} Training light curves $\textbf{x}_i$ undergo 75\% random masking and are processed by a 1D U-Net consisting of an encoder $f_\phi$ and decoder $g_\psi$. The model is trained end-to-end to reconstruct the original unmasked light curves via masked reconstruction loss, learning multi-scale representations. \textbf{(b) Embedding extraction:} After pretraining is complete, each light curve $\textbf{x}_i$ from the simulation bank is passed through the pretrained (frozen) encoder $f_\phi$ to extract 768-dimensional embeddings $\textbf{z}_i$, which are stored with their corresponding parameters as $\{\textbf{z}_i, \boldsymbol{\theta}_i\}$ pairs in a database. \textbf{(c) k-NN retrieval:} For a query observation $\textbf{x}^*$, we encode to $\textbf{z}^* = f_\phi(\textbf{x}^*)$ and perform k-NN retrieval in the learned embedding space by computing cosine similarity $\cos(\textbf{z}_i, \textbf{z}^*)$, retrieving the top-$K=4{,}000$ most similar samples to form an initial empirical posterior $\hat{p}_0(\boldsymbol{\theta} \mid \textbf{x}^*)$ that preserves multimodal parameter structure. \textbf{(d) Local refinement:} Starting from the initial posterior $\hat{p}_0(\boldsymbol{\theta} \mid \textbf{x}^*)$, local hill climbing with forward simulator $F(\boldsymbol{\theta})$ refines each retrieved mode by minimizing Poisson negative log-likelihood $\mathcal{L}(\boldsymbol{\theta}) = F(\boldsymbol{\theta} - \textbf{x}^* \log F(\boldsymbol{\theta})$, producing the refined posterior $\hat{p}_{\textit{ref}}(\boldsymbol{\theta} \mid \textbf{x}^*)$ with increased density at high-likelihood regions.} 
  \label{fig:method}
  % \vskip -0.1in
\end{figure*}

\renewcommand{\labelenumi}{(\alph{enumi})}
\begin{enumerate}
  \item \textbf{Pretrain masked U-Net.} Train a 1D U-Net adapted from \cite{ronneberger2015} to reconstruct masked light curves, yielding an encoder that provides meaningful multi-scale features for retrieval.
  \item \textbf{Build an embedding database.} For each simulated light curve $\mathbf{x}_i$ in the simulation bank, compute a \emph{multi-level} embedding by passing the light curve through the frozen encoder of the pretrained U-Net.
  \item \textbf{Retrieve neighbors for a query observation.} For an observed light curve $\mathbf{x}^*$, used to query the architecture, compute its embedding using the frozen encoder and retrieve the top-$K$ neighbors by cosine similarity in embedding space; the corresponding parameters $\{\boldsymbol{\theta}_k\}$ provide a rough initial empirical approximation to the posterior.

 \item \textbf{Hill-climbing refinement.} Starting from the retrieved parameters $\{\boldsymbol{\theta}_k\}$, minimize the Poisson negative log-likelihood (NLL) using an adaptive $P$-dimensional sign-search hill-climbing procedure. At each iteration, we evaluate the $2^P$ $\pm$ perturbation patterns, accept the best improvement, and adapt the step size until a target NLL is reached or progress plateaus.
\end{enumerate}

\section{Motivation and Rationale for Retrieval and Local Refinement in Posterior Estimation}

Our approach decomposes posterior inference into two complementary stages: a global retrieval stage that constructs an empirical approximation of the posterior (Figure \ref{fig:method}(a–c)), followed by a simulator-guided local refinement stage (Figure \ref{fig:method}(d)) that sharpens the posterior density locally.

Let $\boldsymbol{\theta} \in \mathbb{R}^P$ denote the $P$ physical parameters of an observation, and let $\mathbf{x}^\ast$ be an observation. Our goal is to characterize the posterior distribution
\begin{equation}
p(\boldsymbol{\theta} \mid \mathbf{x}^\ast) \propto p(\boldsymbol{\theta})\,p(\mathbf{x}^\ast \mid \boldsymbol{\theta}),
\end{equation}
which is challenging due to parameter degeneracies in observation: distinct parameter configurations can generate nearly indistinguishable observations. As a result, the posterior distribution is spread over extended, possibly disconnected regions of parameter space rather than concentrated near isolated modes. While Markov chain Monte Carlo (MCMC) methods can, in principle, explore such posteriors accurately, their computational cost can range from days to months for a single observation, particularly when the posterior geometry is highly non-trivial \cite{greg2025}.

\paragraph{Retrieval approximates the global posterior support}
Let $f_\phi : \mathcal{X} \rightarrow \mathbb{R}^m$ be a learned encoder of a masked U-Net model trained via masked observation reconstruction, where $\mathcal{X}$ denotes the space of observations and $\mathbb{R}^m$ an $m$-dimensional latent embedding space. For any observation $\mathbf{x} \in \mathcal{X}$, we define its embedding as $\mathbf{z} = f_\phi(\mathbf{x})$.  

Given a simulation bank $\{(\boldsymbol{\theta}_i, \mathbf{x}_i)\}_{i=1}^N$ of size $N$, where $\boldsymbol{\theta}_i$ denotes a parameter configuration and $\mathbf{x}_i$ the corresponding light curve simulated with that parameter configuration, we first embed each simulated light curve as $\mathbf{z}_i = f_\phi(\mathbf{x}_i)$. For an observed light curve $\mathbf{x}^\ast$ with embedding $\mathbf{z}^\ast = f_\phi(\mathbf{x}^\ast)$, retrieval is performed in the latent space by selecting the $K$ simulated light curves whose embeddings $\{\mathbf{z}_i\}$ are closest to $\mathbf{z}^\ast$ under a chosen similarity metric. The corresponding parameter configurations $\{\boldsymbol{\theta}_i\}$ are then retained, inducing the empirical distribution
\begin{equation}
\hat{p}_0(\boldsymbol{\theta} \mid \textbf{x}^\ast)
\;=\;
\frac{1}{K}
\sum_{i \in \mathcal{I}_K(\textbf{x}^\ast)}
\delta(\boldsymbol{\theta} - \boldsymbol{\theta}_i),
\end{equation}
where $\mathcal{I}_K(\textbf{x}^\ast)$ denotes the index set of the retrieved simulations, $K$ is the number of nearest-neighbor simulations retained, and $\delta(\cdot)$ denotes the Dirac delta distribution (i.e., a point mass located at $\boldsymbol{\theta}_i$). This distribution forms a uniform mixture over parameter values whose simulated light curves are most similar to the query observation $\textbf{x}^\ast$ under the learned representation. Thus, $\hat{p}_0(\boldsymbol{\theta}\mid \textbf{x}^\ast)$ should be interpreted not as an estimate of posterior density, but as an empirical approximation to regions of parameter space supported by the observation. This step is conceptually aligned with approximate Bayesian computation, where inference is performed using informative low-dimensional summary statistics \cite{biau2015new}. Crucially, because retrieval operates globally over a simulation bank, it naturally preserves parameter degeneracies in the empirical posterior by retaining multiple, distinct configurations that explain the data equally well. 

\paragraph{Local refinement sharpens the posterior mass}
While retrieval identifies plausible regions of parameter space, the retrieved samples are not guaranteed to be locally optimal with respect to the likelihood of the observed light curve. Unlike sampling-based approaches such as MCMC, which rely on rejecting and resampling candidates, our method improves the fidelity of each retrieved sample. Specifically, we apply local hill climbing initialized from each $\boldsymbol{\theta}_k \in \mathcal{I}_K(\textbf{x}^\ast)$ by minimizing 
a Poisson negative log-likelihood (NLL)
\begin{equation}
\mathcal{L}_{\text{Poisson}}(\boldsymbol{\theta}_k)
= \frac{1}{T}\sum_{t=1}^{T}
\Big(
F_t(\boldsymbol{\theta}_k) - x^*_t \log\!\big(F_t(\boldsymbol{\theta}_k)+\varepsilon\big)
\Big)
\end{equation}
where \(\varepsilon\) is a tiny pseudocount added inside \(\log\big(F_t(\boldsymbol{\theta}_k)+\varepsilon\big)\) to avoid \(\log0\) and stabilize the Poisson NLL and must be \(> 0\). \(\mathcal{L}_{\text{Poisson}}\) is the mean Poisson NLL between the simulator output \(F(\boldsymbol{\theta}_k)\) and the query observation \(\mathbf{x}^*\) and \(T\) is the number of phase bins in the observation.

Updates are proposed within a restricted neighborhood-- initially modifying single parameters and subsequently allowing larger parameter subsets and are accepted only if they reduce $\mathcal{L}_{\text{Poisson}}(\boldsymbol{\theta}_k)$. This produces a sequence $\{\boldsymbol{\theta}_k^{(n)}\}$ satisfying

\begin{equation}
\mathcal{L}_{\text{Poisson}}(\boldsymbol{\theta}_k^{(n+1)}) \le \mathcal{L}_{\text{Poisson}}(\boldsymbol{\theta}_k^{(n)})
\end{equation}
which corresponds to a monotonic non-decrease of the likelihood $p(\textbf{x}^\ast \mid \boldsymbol{\theta}_k^{(n)})$. Importantly, this refinement remains strictly local: each trajectory improves the fit within the basin of attraction defined by its retrieved initialization. As a result, local optimization sharpens posterior mass within each region of support without collapsing distinct modes or erasing parameter degeneracies captured by the retrieval stage.

Because the embeddings of simulated observations in the simulation bank are computed offline for the retrieval task, and because local refinement guided by the forward simulator is computationally efficient, the proposed retrieval-based approach with local refinement is comparatively tractable, often completing within minutes, relative to traditional MCMC methods, which may require of the order of days of computation using large-scale CPU resources to generate posteriors for a single observation. While the method does not aim to produce exact posterior samples or offer asymptotic guarantees, it provides a practical approximation that retains multimodality and parameter degeneracies in settings where exhaustive Bayesian sampling is computationally challenging.

\section{Masked U-Net for Light Curve Representation Learning}
\label{sec:masked_unet}
We pretrain a 1D U-Net autoencoder adapted from \cite{ronneberger2015} to learn multi-scale representations of pulsar light curves through masked reconstruction. The encoder produces latent features at multiple resolutions, which we subsequently leverage for similarity-based retrieval. More details about the U-Net architecture are presented in Appendix~\ref{sec:unet_appendix}.  

We set the base channel width $\texttt{base\_filters}=64$, which determines the number of filters at each encoder level following a doubling pattern: $64 \to 128 \to 256 \to 512$ channels across the three encoder blocks and bottleneck.

\paragraph{Multi-scale feature hierarchy.} The encoder progressively expands channel capacity ($1 \to 64 \to 128 \to 256 \to 512$ for $\texttt{base\_filters}=64$) while reducing temporal resolution ($64 \to 32 \to 16 \to 8$), yielding representations with increasingly large receptive fields.
Because each encoder block contains two convolutional layers and each downsampling step doubles the effective stride, the receptive field grows rapidly with depth.
For the architecture used here (three encoder blocks followed by a two-layer bottleneck), the effective receptive field at the bottleneck spans approximately 68 phase bins, exceeding the input length of $T=64$.
As a result, bottleneck features integrate information from essentially the entire light curve, capturing global pulse-profile structure, while intermediate features retain higher temporal resolution and encode more localized morphology.

\subsection{Masked Reconstruction Pretraining}

\paragraph{Random masking.}
For each simulated light curve used in training, $\textbf{x}$, we randomly mask a fraction $\rho=0.75$ of phase bins. Given normalized input $\tilde{\mathbf{x}} \in \mathbb{R}^{T}$, we sample a random permutation $\pi$ of indices $\{1,\ldots,T\}$ and retain only $T_{\text{keep}} = \lfloor(1-\rho)T\rfloor = 16$ positions. The binary mask $\mathbf{m} \in \{0,1\}^T$ is:
\begin{equation}
\label{eq:mask_generation}
m_t = \begin{cases}
0 & \text{if } t \in \{\pi(1), \ldots, \pi(T_{\text{keep}})\} \\
1 & \text{otherwise}
\end{cases}.
\end{equation}
Masked positions are zeroed in the input:
\begin{equation}
\label{eq:masked_input}
\tilde{\mathbf{x}}^{\text{mask}} = (1 - m_t) \cdot \tilde{\textbf{x}}.
\end{equation}
\paragraph{Masking ratio.}
During training, we apply per-sample random masking with a mask fraction of $0.75$,
so that $75\%$ of phase bins are masked and reconstructed. During inference and embedding extraction, we disable masking by setting
$\texttt{mask\_ratio}=0.0$, ensuring that the encoder processes the complete light curve
when extracting bottleneck and multi-level latent representations.

\paragraph{Reconstruction objective.}
The U-Net decoder $g_\psi$ reconstructs the \emph{original} unmasked input from the bottleneck and skip connections. Let $\hat{\mathbf{x}} = g_\psi(f_\phi(\tilde{\mathbf{x}}^{\text{mask}}))$ denote the reconstruction. The loss is computed \emph{only on masked positions}:
\begin{equation}
\label{eq:masked_loss}
\mathcal{L}(\phi, \psi) = \mathbb{E}_{(\mathbf{x}, \mathbf{m}) \sim \mathcal{D}} \left[\frac{\sum_{t=1}^T m_t (\hat{x}_t - \tilde{x}_t)^2}{\sum_{t=1}^T m_t}\right],
\end{equation}
where the expectation is over light curves $\mathbf{x}$ from training distribution $\mathcal{D}$ and random masks $\mathbf{m}$. By penalizing reconstruction error only where the model lacks direct observations, this objective encourages the encoder to learn contextual representations that capture phase-resolved structure necessary for inferring missing flux values.

\paragraph{Training procedure.}
We pretrain on $N=10M$ simulated light curves using AdamW with learning rate $\eta_0=10^{-5}$, linearly decayed to zero over 200 epochs. 

\section{Embedding Construction and Retrieval}
\label{sec:embeddings}

After pretraining, the encoder $f_\phi$ %of encoder 
produces multi-scale features from a light curve that can be aggregated into fixed-dimensional embeddings for similarity-based retrieval. We compare two aggregation strategies: single-level (bottleneck only) and multi-level (intermediate + bottleneck).

\subsection{Single-Level Embeddings}
\label{subsec:single_level}

The bottleneck representation $\mathbf{b} \in \mathbb{R}^{512 \times 8}$ from Eq.~\eqref{eq:encoder_hierarchy} (Appendix) captures global context through its large receptive field. We obtain a fixed-length embedding via temporal pooling \cite{lin2013network}:
\begin{equation}
\label{eq:single_embed}
\mathbf{z}^{\text{single}} = \mathrm{GAP}(\mathbf{b}) = \frac{1}{8}\sum_{t=1}^{8} \mathbf{b}_{:,t} \in \mathbb{R}^{512},
\end{equation}
where $\mathrm{GAP}$ denotes global average pooling over the temporal dimension. This produces a permutation-invariant descriptor emphasizing coarse periodic structure.

\subsection{Multi-Level Embeddings}
\label{subsec:multi_level}

Single-level embeddings primarily encode global shape through deep features with large receptive fields. To additionally capture local pulse morphology, we fuse intermediate features with the bottleneck.

\paragraph{Intermediate feature extraction.}
At encoder level 3, the feature map $\mathbf{e}^{(3)} \in \mathbb{R}^{256 \times 16}$ maintains higher temporal resolution and smaller receptive fields than the bottleneck. We extract this via forward hooks during inference and apply temporal pooling:
\begin{equation}
\label{eq:level3_pool}
\mathbf{f}^{(3)} = \mathrm{GAP}(\mathbf{e}^{(3)}) \in \mathbb{R}^{256}.
\end{equation}

\paragraph{Hierarchical concatenation.}
We concatenate the pooled intermediate features with the single-level embedding:
\begin{equation}
\label{eq:multi_embed}
\mathbf{z}^{\text{multi}}_{\text{raw}} = [\mathbf{f}^{(3)} \oplus \mathbf{z}^{\text{single}}] \in \mathbb{R}^{768},
\end{equation}
where $\oplus$ denotes concatenation. This parameter-free fusion preserves information from both hierarchical levels without requiring additional training.

\paragraph{Normalization for cosine retrieval.}
All embeddings are $\ell_2$-normalized to unit length:
\begin{equation}
\label{eq:l2_norm}
\mathbf{z} = \frac{\mathbf{z}_{\text{raw}}}{\|\mathbf{z}_{\text{raw}}\|_2 + \epsilon}, \quad \epsilon = 10^{-8}.
\end{equation}
Under this normalization, cosine similarity equals inner product: $\cos(\mathbf{z}_i, \mathbf{z}_j) = \mathbf{z}_i^\top \mathbf{z}_j$, enabling efficient retrieval via matrix multiplication.

\subsection{k-Nearest Neighbor (k-NN) Retrieval}
\label{subsec:knn_retrieval}

Given a query light curve $\mathbf{x^*}$, we use two-stage retrieval. First, we compute cosine similarity in embedding space:
\begin{equation}
\label{eq:cosine_sim}
s_i = \mathbf{z^*}^\top \mathbf{z}_i, \quad \text{where } \mathbf{z^*} = f_\phi(\mathbf{x^*}),
\end{equation}
and retrieve the top 20,000 most similar light curves from the database. Second, we re-rank these 20,000 candidates using the MdNSE metric \cite{greg2025}--a scale-invariant measure in unnormalized signal space (Appendix~\ref{app:mdnse})--selecting the final $K=4{,}000$ neighbors with lowest reconstruction error.

This hybrid approach balances efficiency (cosine similarity over 10M embeddings) with accuracy (MdNSE refinement in physical flux space). The parameters corresponding to the top-$K$ retrieved light curves form the initial empirical posterior $\hat{p}_0(\boldsymbol{\theta} \mid \mathbf{x^*})$ as described in Section 3.

\section{Hill Climbing Optimization for Refined Posterior Prediction}
We refine the retrieval-initialized empirical posterior using a forward model \cite{greg2025}–guided local search \cite{tovey1985} that climbs toward higher likelihood regions while remaining simulator-consistent. The refinement operates in parameter space 
\(\boldsymbol{\theta} \in \mathbb{R}^{11}\) and on light-curve outputs of length \(T=64\).

\subsection{Objective functions}

Given target (observed) light curve $\mathbf{x}^* \in \mathbb{R}^{64}$ and simulator 
\(F(\boldsymbol{\theta})\), we optimize a Poisson negative log-likelihood (NLL)
\begin{equation}
\mathcal{L}_{\text{Poisson}}(\boldsymbol{\theta})
= \frac{1}{T}\sum_{t=1}^{T}
\Big(
F_t(\boldsymbol{\theta}) - x^*_t \log\!\big(F_t(\boldsymbol{\theta})+\varepsilon\big)
\Big)
\end{equation}
where \(\varepsilon>0\), \(\mathcal{L}_{\text{Poisson}}\) is the mean Poisson NLL between the simulator output \(F(\theta)\) and the observed light curve \(\mathbf{x}^*\) and \(T\) is the number of phase bins in the light curve.

\subsection{From 1-D to \(P\)-D hill climbing}
A simple baseline updates one parameter at a time: at each step it proposes 
\(\boldsymbol{\theta}_j \pm \eta \) for a randomly chosen \(j\) and accepts the better move. This is robust but can be slow when parameters are correlated \cite{wright1996direct, kolda2003optimization}.

We generalize to block hill climbing over \(P\) parameters chosen adaptively %each outer iteration, where an outer iteration refers to 
during each full cycle of the block hill-climb update. For a chosen index set \( I = \{ i_1, i_2, \ldots, i_P \} \), we evaluate all \(2^{P}\) signed moves of magnitude \(\eta\) along those axes.

Let the step magnitudes be $\Delta=\big(\Delta_1,\ldots,\Delta_P\big)\in\mathbb{R}_+^P$.
We enumerate all $2^P$ sign patterns
\[
\sigma \in \{-1,+1\}^P,
\]
and form the candidate set, we pick the best joint move by discrete search:
\[
\sigma^\star \;=\; \arg\min_{\sigma \in \{-1,+1\}^P} \; \mathcal{L}\!\left(\boldsymbol{\theta}^{(\sigma)}\right),
\quad
\boldsymbol{\theta}^{+} \;=\; \boldsymbol{\theta}^{(\sigma^\star)}.
\]

\subsection{Adaptive step control}
After each block, update the step magnitudes by
\[
\Delta \leftarrow
\begin{cases}
\gamma_{\text{grow}}\;\Delta, & \text{if } \mathcal{L}(\boldsymbol{\theta}^{+}) < \mathcal{L}(\boldsymbol{\theta}) - \varepsilon,\\[4pt]
\gamma_{\text{shrink}}\;\Delta, & \text{otherwise},
\end{cases}
\]
with $\,\gamma_{\text{grow}}>1$, $\,\gamma_{\text{shrink}}\in(0,1)$, and tolerance $\varepsilon>0$.
Optionally stop when no improvement occurs for $p$ consecutive blocks (plateau) or when
$\min_j \Delta_j < \Delta_{\min}$.

\subsection{Block Refinement}
At the core is a repeated block refinement loop. Given a seed \(\boldsymbol{\theta} ^{(0)}\), we perform a sequence of blocks; each block executes T inner trial steps. In a trial, we randomly select an index set \[\tilde{\boldsymbol{\theta}}_i \leftarrow \boldsymbol{\theta}_i + s\,\sigma_i,\quad i \in I,\ \quad \sigma_i \in \{-1,+1\},\] with clamping to parameter bounds. We accept the proposal if it reduces \(\mathcal{L}\). Within each block we anneal the step size \(s\) if the acceptance rate falls below a target band.

\paragraph{Stopping.}We maintain the best value per seed, \(\mathcal{L}
^*\), the block gain \(\Delta b\) (improvement after block b), and a patience counter. Refinement stops when (i) \(\Delta b < \epsilon\) for P consecutive blocks (plateau), or (ii) a target NLL \(\tau\) is reached (e.g., the best refined baseline), or (iii) a maximum number of blocks is exceeded.

\section{Experiments and Results}
We evaluate our framework on PSR J0030+0451, a millisecond pulsar observed by NICER \citep{gendreau2016nicer}, using a 10M light curve (LC) simulation database \citep{greg2025}. Each simulation is parameterized by an 11-dimensional vector $\boldsymbol{\theta} = (x_D, y_D, z_D, \alpha_D, \phi_D,\; x_Q, y_Q, z_Q, \alpha_Q, \phi_Q,\; B_Q/B_D)$, where $(x_D,y_D,z_D)$ and $(x_Q,y_Q,z_Q)$ are the Cartesian offsets of the dipole and quadrupole components relative to the neutron-star center, $(\alpha_D,\phi_D)$ and $(\alpha_Q,\phi_Q)$ are their corresponding orientation angles (inclination and azimuth) and $B_Q/B_D$ is the quadrupole-to-dipole surface field-strength ratio.

\subsection{Embedding Space and Retrieval Performance}
We embedded the 10M light curve database using the pretrained masked U-Net encoder, extracting 768-dimensional multi-level embeddings by concatenating pooled encoder level-3 (256-dim; Eq.~\eqref{eq:level3_pool}) and bottleneck (512-dim; Eq.~\eqref{eq:single_embed}) features. This representation captures both local phase-resolved structure and global pulse morphology, enabling cosine-similarity retrieval in latent space (Eq.~\eqref{eq:cosine_sim}). The retrieval completes in $\sim$17.6 seconds on a CPU, as reported in Table~\ref{tab:timing}.

For the observed PSR J0030+0451 light curve, we compared three retrieval strategies: (i) cosine similarity on raw z-scored light curves, (ii) single-level bottleneck embeddings, and (iii) multi-level embeddings. Retrieval quality is assessed using the MdNSE metric introduced by \citet{greg2025} (Eq.~\eqref{eq:quality_metric}). The qualitative retrieval overlays are reported in Appendix~\ref{app:ablation}.
% achieved the lowest MdNSE 
Multi-level embeddings produced retrieved curves that most faithfully match phase-dependent structure, including peak width, asymmetry, and baseline morphology. Bottleneck-only embeddings showed modest degradation, while raw-space cosine similarity yielded frequent peak misalignment and baseline mismatch.

\subsection{Hill-Climbing Refinement Results}
Starting from $K{=}4{,}000$ retrieved candidates (\ref{sec:justify_K_4000}), we applied adaptive hill-climbing refinement with $P$ jointly-perturbed parameters at each iteration. Table~\ref{tab:refinement} shows systematic convergence toward the MCMC reference posterior as $P$ increases from 1 to 5.

% TABLE 1: Hill-Climbing Performance
\begin{table}[t]
\caption{Hill-climbing refinement systematically converges to MCMC posterior. Each row shows results after refining all 4,000 seeds with the specified number of jointly-perturbed parameters $P$.}
\label{tab:refinement}
\centering
\small
\begin{tabular}{lccc}
\toprule
\textbf{P} & \textbf{Mean NLL} & \textbf{$\Delta$ from Initial} & \textbf{$\Delta$ from MCMC} \\
% & & & Initial \\
\midrule
Initial & $-41{,}513.00$ & --- & $-612.14$ \\
1 & $-42{,}089.93$ & $+576.93$ & $-35.21$ \\
2 & $-42{,}122.51$ & $+609.51$ & $-2.63$ \\
3 & $-42{,}123.58$ & $+610.58$ & $-1.56$ \\
4 & $-42{,}124.07$ & $+611.07$ & $-1.07$ \\
5 & $-42{,}124.30$ & $+611.30$ & $-0.84$ \\
\midrule
MCMC & $-42{,}125.14$ & --- & --- \\
\bottomrule
\end{tabular}
\end{table}

Refinement with $P{=}5$ achieves within 0.84 NLL units of nested sampling MCMC, with most improvement occurring by $P{=}2$. Beyond $P{=}3$, gains plateau with diminishing returns: the improvement from $P{=}4$ to $P{=}5$ is only $+0.23$ NLL units. This indicates that jointly perturbing 5 parameters is sufficient to navigate the complex, degenerate posterior landscape of the 11-dimensional pulsar parameter space.  

\begin{figure*}[h!]
  \centering
  \includegraphics[width=0.9755\linewidth]{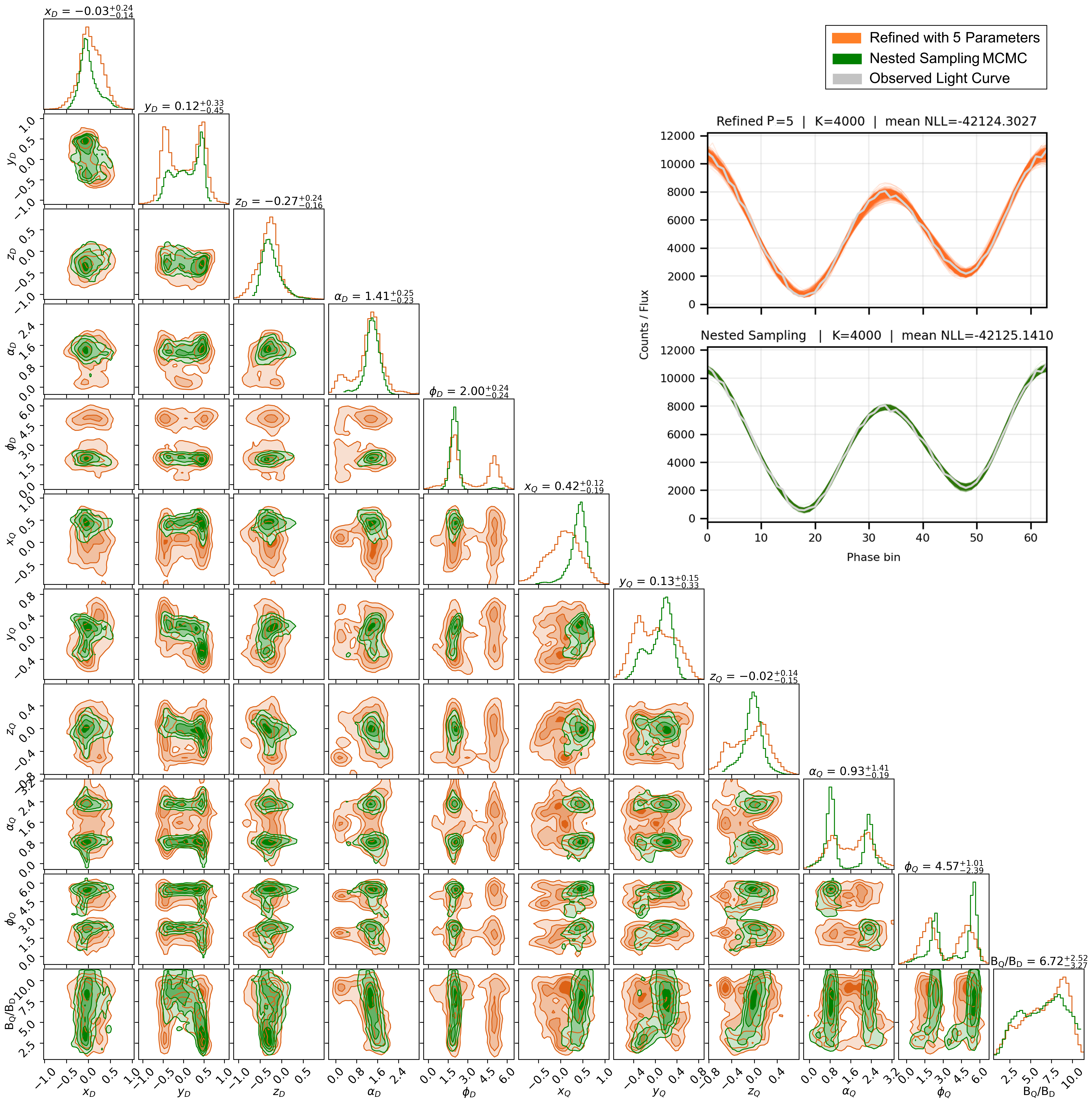}
  \caption{\textbf{Left}: Posterior distributions of pulsar parameters comparing results from (i) the embedding retrieval and hill climbing refinement over 5 parameters approach in the work (in orange) and (ii) a traditional MCMC informed by a nested sampling approach (in green). \textbf{Right}: Predicted light curves corresponding to parameter posterior configurations from this work (orange) compared to those from the baseline MCMC pipeline (green). The predicted posteriors align well with the observed light curve and are in agreement with the predictions from MCMC.}
  \label{fig:observed_lc}
\end{figure*}

Figure~\ref{fig:observed_lc} shows refined posterior distributions for parameters of PSR J0030+0451. We map them to corresponding predicted light curves using out light curve simulator. We find that our predicted parameter posteriors match those from MCMC across all parameters while preserving multimodal structure in degenerate dimensions. The parameter posteriors map to the same region of the light-curve space as the observed light curve and align with predicted light curves from the MCMC inference approach used as a baseline in this work. The quadrupole parameters $\{x_Q, y_Q, z_Q, \alpha_Q, \phi_Q, B_Q/B_D\}$ exhibit particularly strong agreement with MCMC, while the quadrupole orientation angles $\{\alpha_Q, \phi_Q\}$ preserve the expected multimodal structure arising from parameter degeneracies. Notably, for $\alpha_D$ and $\phi_D$, the refined posterior captures multiple distinct modes that reflect the inherent symmetries in the magnetic field geometry. This demonstrates that local refinement successfully sharpens posterior density within each mode without collapsing the degenerate structure captured during retrieval.

\subsection{Why is the Retrieval Size K=4,000?}
\label{sec:justify_K_4000}
We empirically determine the optimal number of retrieved neighbors $K$ by measuring convergence of the approximate posterior as we increase $K$ s.t. $K_0 < K_1 < \cdots <K_\text{max}$. For each parameter ${\theta}_i$, we compute the KL divergence between the empirical distribution formed by the top-$K_k$ retrieved samples and the top-$K_\text{max}$ retrieved samples.

\begin{equation}
    D_{KL}^{(i)}(K) = \sum_{\text{bins}} p_{\text{retrieved}}^{(i)}({\theta}_i; K_\text{max}) \log\frac{p_{\text{retrieved}}^{(i)}({\theta}_i; K_\text{max})}{p_{\text{retrieved}}^{(i)}({\theta}_i; K)},
\end{equation}

where both distributions are discretized using histogram binning with consistent bin edges across all $K$ values.

\begin{figure}[h!]
  \centering
  \includegraphics[width=\linewidth]{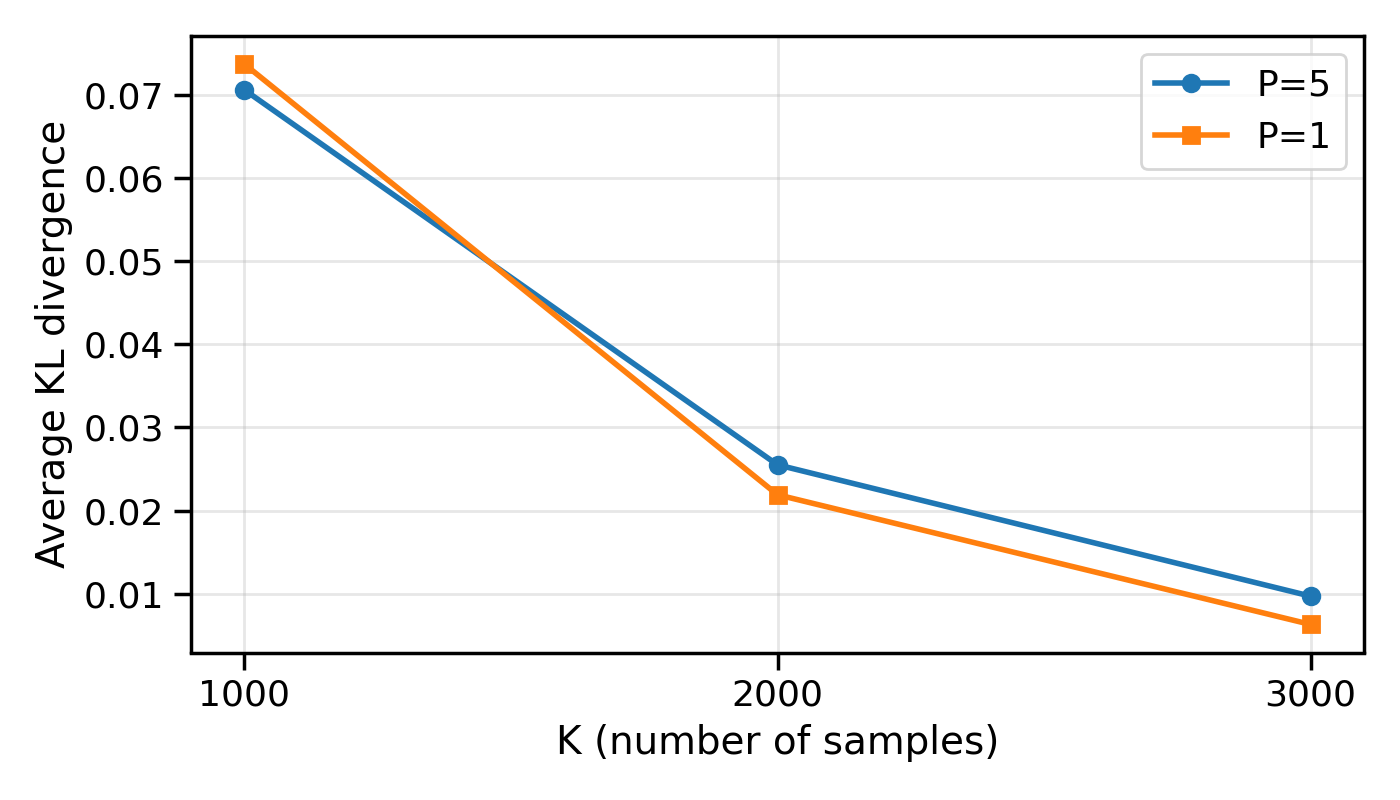}
  \caption{Average KL divergence versus retrieval size $K$ for $P{=}1$ (orange) and $P{=}5$ (blue) refinement. KL divergence decreases monotonically and levels off at $K\approx 3000$ (KL $\approx 0.01$), justifying the use of $K{=}4{,}000$. } 
  \label{fig:KL}
\end{figure}

In Figure~\ref{fig:KL}, the average KL divergence monotonically decreases with the increase of $K$, where we measure against $K_\text{max} = 4,000$. Both refinement strategies $(P=1$ and $P=5)$ converge so that the estimated posteriors for $K=3,000$ and $K=4,000$ are nearly identical.

Based on this convergence, we select $K=4000$ for all experiments to ensure the retrieval stage captures the full support of the posterior while maintaining computational efficiency.

\subsection{Computational Performance}

Table~\ref{tab:timing} summarizes end-to-end computational costs. Our method achieves \textbf{120$\times$ speedup} over MCMC (24 hours $\rightarrow$ 12 minutes) while maintaining posterior fidelity.

% TABLE 2: Computational Cost
\begin{table}[t]
\caption{End-to-end computational cost breakdown for posterior inference.}
\label{tab:timing}
\centering
\small
\begin{tabular}{llr}
\toprule
\textbf{Component} & \textbf{Hardware} & \textbf{Time} \\
\midrule
\textit{Offline (one-time)} \\
\quad Embed 10M LCs & RTX A5000 GPU & $\sim$6.6 min \\
\midrule
\textit{Online (per query)} \\
\quad Query embedding & Apple M-series & $<$1 sec \\
\quad k-NN retrieval & Apple M-series & 17.6 sec \\
\quad Refinement ($P{=}1-5$) & 4,000 CPUs  & 11.8 min \\
\quad \textbf{Total online} & & \textbf{$\sim$12 min} \\
\midrule
\textit{Baseline} \\
\quad MCMC & 4,000 CPUs & \textbf{24 hours} \\
\bottomrule
\end{tabular}
\end{table}

\textbf{Key insights:}
\textbf{(i)} Offline preprocessing (6.6 min GPU for 10M light curves at 25,300 LC/s) is amortized across all queries---a 120$\times$ speedup over CPU (14 hours).
\textbf{(ii)} Online inference (~12 min) breaks down as retrieval (~17.6s) plus refinement (~11.8) min).
\textbf{(iii)} On GPUs processing of all 4,000 refinement seeds in parallel take 66.5 seconds per seed, compared to 363 seconds per seed on CPU (5.5$\times$ speedup).
\textbf{(iv)} Retrieval scales logarithmically with database size, enabling real-time inference even for billion-scale simulation banks.

\subsection{Generalization to Gamma-Ray Spectroscopy}

To assess domain transferability, we applied the framework to Ba-133 photopeak fitting—a 4-parameter inference problem in nuclear spectroscopy. The task is to estimate amplitude $a$, peak center $x_c$, width $w$, and background $y_0$ from a Gaussian-plus-background model $f(E; \boldsymbol{\theta}) = a \exp(-(E - x_c)^2 / 2w^2) + y_0$, where $E \in [156, 164]$ keV is photon energy.  We generated 10M synthetic photopeaks by sampling parameters from priors centered on an observed Ba-133 spectrum and applying Poisson observation noise. Each spectrum was resampled to $40$ uniformly-spaced energy bins. Training the same U-Net architecture (base\_filters=64, 768-dim embeddings, 20 epochs) on this 10M synthetic spectra top-$K=4{,}000$ retrieval achieved cosine similarity $>0.999$ with observed photopeaks (Figure~\ref{fig:gamma_retrieval} in Appendix). Hill-climbing refinement (Poisson NLL, $P=3$) reduced mean NLL from $-9227.6425$ to $-9228.7829$, matching the MCMC reference. Results are included in Appendix~\ref{app:grs}.

\section{Conclusions}
We introduced an accelerated framework for posterior inference from pulsar light curves that avoids computationally expensive MCMC sampling. We pretrain a U-Net model to reconstruct masked light curves, yielding an encoder that learns a meaningful multi-scale embedding space for pulsar light-curve representations. Given a query light curve, our approach retrieves simulated light curves with similar characteristics from a simulation bank by measuring similarity in the learned embedding space. The parameter sets associated with the retrieved light curves define an empirical approximation to the posterior distribution, naturally capturing multimodality and parameter degeneracies present in the data. We then locally refine each retrieved region of parameter space using hill-climbing optimization guided by a forward simulator, sharpening posterior mass within each region of support while preserving distinct modes and degeneracies identified during the retrieval stage. When applied to NICER observations of PSR J0030+0451, our approach achieves a $120\times$ reduction in inference time while maintaining quantitative agreement across all inferred parameters. The method offers an empirically effective approximation for likelihood-free inference when exhaustive Bayesian sampling is computationally impractical. In addition to pulsar modeling, results on a gamma-ray spectrum inference task indicate that the framework generalizes to other physics-based inference problems involving complex posterior structure.

\section*{Acknowledgements}
This work was supported by the NASA theoretical and computational astrophysics network project with grant number 22-TCAN22-0027. A.B.S., D.O., and S.D. were supported by the Laboratory Directed Research and Development (LDRD) program of LANL through its Information Science \& Technology Institute under project number 20258368CT-IST. This research used resources provided by the Los Alamos National Laboratory Institutional Computing Program, which is supported by the U.S. Department of Energy National Nuclear Security Administration under Contract No. 89233218CNA000001.

\bibliography{arxiv_paper}
\bibliographystyle{arxiv2026}

% APPENDIX
\newpage
\appendix
\onecolumn
\section{U-Net Architecture}
\label{sec:unet_appendix}

\subsection{Encoder Architecture}
\label{subsec:unet_encoder}

Let $\mathbf{x} \in \mathbb{R}^{T}$ denote a light curve with $T$ phase bins. After z-score normalization and reshaping to channel-first format $\tilde{\mathbf{x}} \in \mathbb{R}^{1 \times T}$, the encoder $f_\phi: \mathbb{R}^{1 \times T} \to \mathbb{R}^{C_b \times T_b}$ applies $L=3$ hierarchical blocks with progressive downsampling:

\begin{equation}
\label{eq:encoder_hierarchy}
\begin{aligned}
\mathbf{e}^{(1)} &= \mathrm{Enc}_1(\tilde{\mathbf{x}}^{\text{mask}}) &&\in \mathbb{R}^{64 \times 64}, \\
\mathbf{e}^{(2)} &= \mathrm{Enc}_2(\mathrm{Pool}(\mathbf{e}^{(1)})) &&\in \mathbb{R}^{128 \times 32}, \\
\mathbf{e}^{(3)} &= \mathrm{Enc}_3(\mathrm{Pool}(\mathbf{e}^{(2)})) &&\in \mathbb{R}^{256 \times 16}, \\
\mathbf{b} &= \mathrm{Enc}_b(\mathrm{Pool}(\mathbf{e}^{(3)})) &&\in \mathbb{R}^{512 \times 8},
\end{aligned}
\end{equation}
where $\mathbf{e}^{(\ell)}$ denotes the $\ell$-th encoder feature map, $\mathbf{b}$ is the bottleneck representation, and $\mathrm{Pool}(\cdot)$ is max-pooling with stride 2.

\paragraph{Convolutional blocks.}
Each encoder block $\mathrm{Enc}_\ell$ consists of two 1D convolutional layers with ReLU activations:
\begin{equation}
\label{eq:conv_block}
\mathrm{Enc}_\ell(\mathbf{u}) = \sigma(\mathbf{W}_\ell^{(2)} \ast \sigma(\mathbf{W}_\ell^{(1)} \ast \mathbf{u})),
\end{equation}
where $\ast$ denotes 1D convolution with kernel size 3, padding 1, and $\sigma(\cdot) = \max(0, \cdot)$ is ReLU activation function. This configuration preserves spatial dimensions within each block while $\mathrm{Pool}$ halves temporal resolution between blocks.

\subsection{Decoder Architecture}
The U-Net decoder $g_\psi$ mirrors the encoder hierarchy via three transpose-convolution upsampling stages with concatenated skip-connections. Let $\mathbf{b}\in\mathbb{R}^{512\times 8}$ be the bottleneck feature and $\mathbf{e}^{(\ell)}$ the encoder feature maps. The decoder computes:

\[
\begin{aligned}
\mathbf{u}^{(3)} &= \mathrm{Up}_3(\mathbf{b}) \in \mathbb{R}^{256 \times 16},\\
\mathbf{d}^{(3)} &= \mathrm{Dec}_3\big([\mathbf{u}^{(3)}\Vert \mathbf{e}^{(3)}]\big) \in \mathbb{R}^{256 \times 16},\\
\mathbf{u}^{(2)} &= \mathrm{Up}_2(\mathbf{d}^{(3)}) \in \mathbb{R}^{128 \times 32},\\
\mathbf{d}^{(2)} &= \mathrm{Dec}_2\big([\mathbf{u}^{(2)}\Vert \mathbf{e}^{(2)}]\big) \in \mathbb{R}^{128 \times 32},\\
\mathbf{u}^{(1)} &= \mathrm{Up}_1(\mathbf{d}^{(2)}) \in \mathbb{R}^{64 \times 64},\\
\mathbf{d}^{(1)} &= \mathrm{Dec}_1\big([\mathbf{u}^{(1)}\Vert \mathbf{e}^{(1)}]\big) \in \mathbb{R}^{64 \times 64},\\
\hat{\mathbf{x}} &= \mathrm{Conv}_{1\times1}\!\big(\mathbf{d}^{(1)}\big) \in \mathbb{R}^{1 \times 64}.
\end{aligned}
\]

Each $\mathrm{Up}_\ell$ upsamples via transposed convolution (kernel size 2, stride 2), halving channel depth while doubling spatial resolution. Concatenation $[\cdot\Vert\cdot]$ along the channel dimension doubles the input channels to each $\mathrm{Dec}_\ell$ block. Each decoder block applies:
\[
\mathrm{Dec}_\ell(\mathbf{z}) = \sigma\!\big(W_\ell^{(2)} \ast \sigma(W_\ell^{(1)} \ast \mathbf{z})\big),
\]
where the first convolution processes concatenated features $(2C_{\text{up}} \to C_{\text{out}})$ and the second refines within $C_{\text{out}}$.

\section{Evaluation Metric for Retrieved Light Curve Quality} 
\label{app:mdnse}

To quantitatively evaluate the quality of retrieved light curves, we employ a median normalized squared error (MdNSE) metric \cite{greg2025} computed in the unnormalized signal space. Given a query light curve $\tilde{\mathbf{x}} \in \mathbb{R}^{T}$ and a retrieved candidate $\tilde{\mathbf{x}}_\textit{retrieved } \in \mathbb{R}^{T}$ (both z-score normalized), we first invert the normalization:
\begin{equation}
\label{eq:unnormalize}
 \mathbf{x} = \sigma  \tilde{\mathbf{x}} + \mu, \quad \mathbf{x}_\textit{retrieved}  = \sigma \tilde{\mathbf{x}}_\textit{retrieved} + \mu,
\end{equation}
where $\mu$ and $\sigma$ are the precomputed mean and standard deviation from the training data set.

\begin{equation}
\label{eq:quality_metric}
\text{MdNSE} = \frac{\sum_{t=1}^{T} (x_{\textit{retrieved} ,t}  - x_{t})^2}{\left(\text{median}(\mathbf{x})\right)^2},
\end{equation}

where the numerator measures the sum of squared pointwise errors in unnormalized space, and the denominator normalizes by the squared median flux of the query light curve, making the metric scale-invariant. Lower MdNSE values indicate better agreement between the retrieved and query light curves. The MdNSE was chosen over $\chi^2$ because the simulated data lack associated error values, and the metric treats all light curves more equally by focusing on shape rather than amplitude \cite{greg2025}.

\section{Ablation Results}
\label{app:ablation}
We ablate the retrieval representation by comparing three similarity spaces: (i) cosine similarity on raw z-scored light curves, (ii) bottleneck-only embeddings (global average pooled U-Net bottleneck), and (iii) multi-level embeddings formed by concatenating pooled encoder level-3 features with pooled bottleneck features. Across multiple query light curves, multi-level embeddings produce the most faithful nearest-neighbor sets, yielding retrieved curves that tightly track the query’s phase-dependent morphology (peak width, asymmetry, and baseline structure), while bottleneck-only embeddings show modestly larger dispersion. In contrast, raw-space cosine similarity is substantially less reliable and often retrieves curves with mismatched peak shapes and off-peak baselines. These results indicate that combining intermediate and bottleneck features is important for preserving both local and global structure in the embedding space, improving retrieval fidelity for posterior initialization. 

\begin{figure*}[t]
\includegraphics[
  width=\columnwidth,
  height=0.9\textheight,
  keepaspectratio
  ]{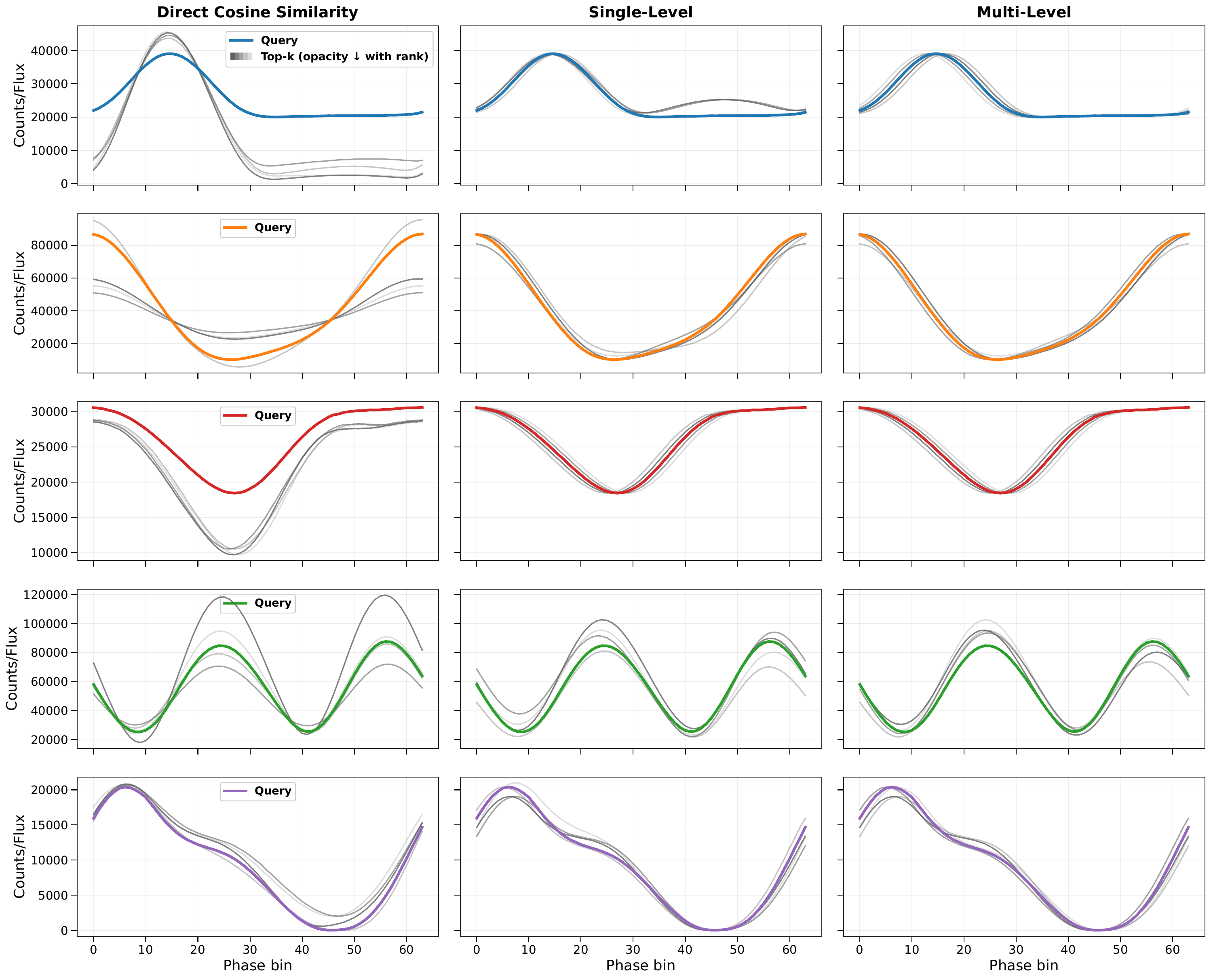}
  \caption{Visual diagnostic of retrieval fidelity. For each query (row), the spread of retrieved curves across ranks reflects how well the similarity space concentrates around the observed morphology; multi-level embeddings show the smallest spread, whereas raw-space cosine similarity exhibits larger deviations in peak and baseline structure.}
  \label{fig:ablation}
\end{figure*}

Figure~\ref{fig:ablation} compares nearest-neighbor retrieval across three similarity spaces: bottleneck-only embeddings obtained via global average pooling of the U-Net bottleneck, multi-level embeddings that fuse encoder level-3 features with bottleneck features, and direct cosine similarity computed on raw, z-scored light curves. For each query (rows), we visualize the observed light curve (colored) together with the top-$K$, where $K=5$, retrieved matches (gray, with opacity decreasing by rank). Across all queries, retrieval in the learned embedding space consistently produces matches that more closely follow the query morphology than direct cosine similarity on raw curves.

Multi-level embeddings yield the tightest retrieval sets, with the top-$K$ curves forming compact envelopes around the query and exhibiting minimal phase-dependent deviations. This improvement arises because the multi-level representation combines complementary information from different depths of the encoder: bottleneck features encode global pulse-profile structure through their large receptive fields, while intermediate encoder features (level 3) retain higher phase resolution and capture localized morphological details such as peak width, asymmetry, and shoulder structure. Encoder level-1 features would be too local and susceptible to high-frequency fluctuations, whereas level-3 provides a “sweet spot”—local enough to preserve fine, phase-resolved structure, yet deep enough to suppress noise and irrelevant variations. In contrast, single-level embeddings rely solely on bottleneck features and therefore emphasize global shape while partially discarding fine-scale phase information, leading to slightly increased variability among retrieved matches.

Direct cosine similarity on raw light curves performs substantially worse, frequently retrieving curves with mismatched peak shapes and off-peak baselines. This behavior reflects the absence of learned invariances in raw phase space, where cosine similarity is overly sensitive to local fluctuations and insufficiently discriminative under strong parameter degeneracies. Overall, these results support the use of multi-level embeddings for retrieval-based posterior initialization, as they more faithfully preserve both global and local structure in pulsar light curves.

\section{Gamma-Ray Spectroscopy: Posterior Visualization and Comparison}
\label{app:grs}

We evaluate the proposed retrieval-based posterior inference framework on a
gamma-ray spectroscopy (GRS) problem involving photopeak fitting for a
Ba-133 spectrum, following the standard formulation used in PyMC3 and
dynesty demonstrations \citep{moghani2025sampling}. 

\subsection{Forward Model}
Each photopeak is modeled as a Gaussian signal superimposed on a constant
background:
\begin{equation}
F(E; \boldsymbol{\theta}) =
a \exp\!\left(-\frac{(E - x_c)^2}{2w^2}\right) + y_0,
\end{equation}
where $\boldsymbol{\theta} = (a, x_c, w, y_0)$ denote the amplitude, peak
center (keV), standard deviation (keV), and background level, respectively.
The energy axis $E$ is discretized into 40 uniformly spaced bins over the
photopeak region.

\subsection{Synthetic Dataset and Retrieval}

Synthetic spectra were generated by sampling parameters from Gaussian priors
centered on an observed Ba-133 photopeak, consistent with the priors used in
the PyMC3 and dynesty reference implementation \cite{moghani2025sampling}. Poisson noise was applied to simulate photon-counting statistics. A total of 10M synthetic spectra were generated and we used the same masked 1D U-Net architecture and training procedure described
in Section~\ref{sec:masked_unet}, without modification.

For a query spectrum, the pretrained encoder produces a fixed-dimensional
embedding, and cosine similarity in embedding space is used to retrieve the
top-$K=4{,}000$ nearest simulated spectra. The corresponding parameter sets
form an empirical posterior approximation over $(a, x_c, w, y_0)$.

\subsection{Results after Refinement}

\begin{figure}[ht]
  \vskip 0.2in
  \begin{center}
    \centerline{\includegraphics[width=\columnwidth]{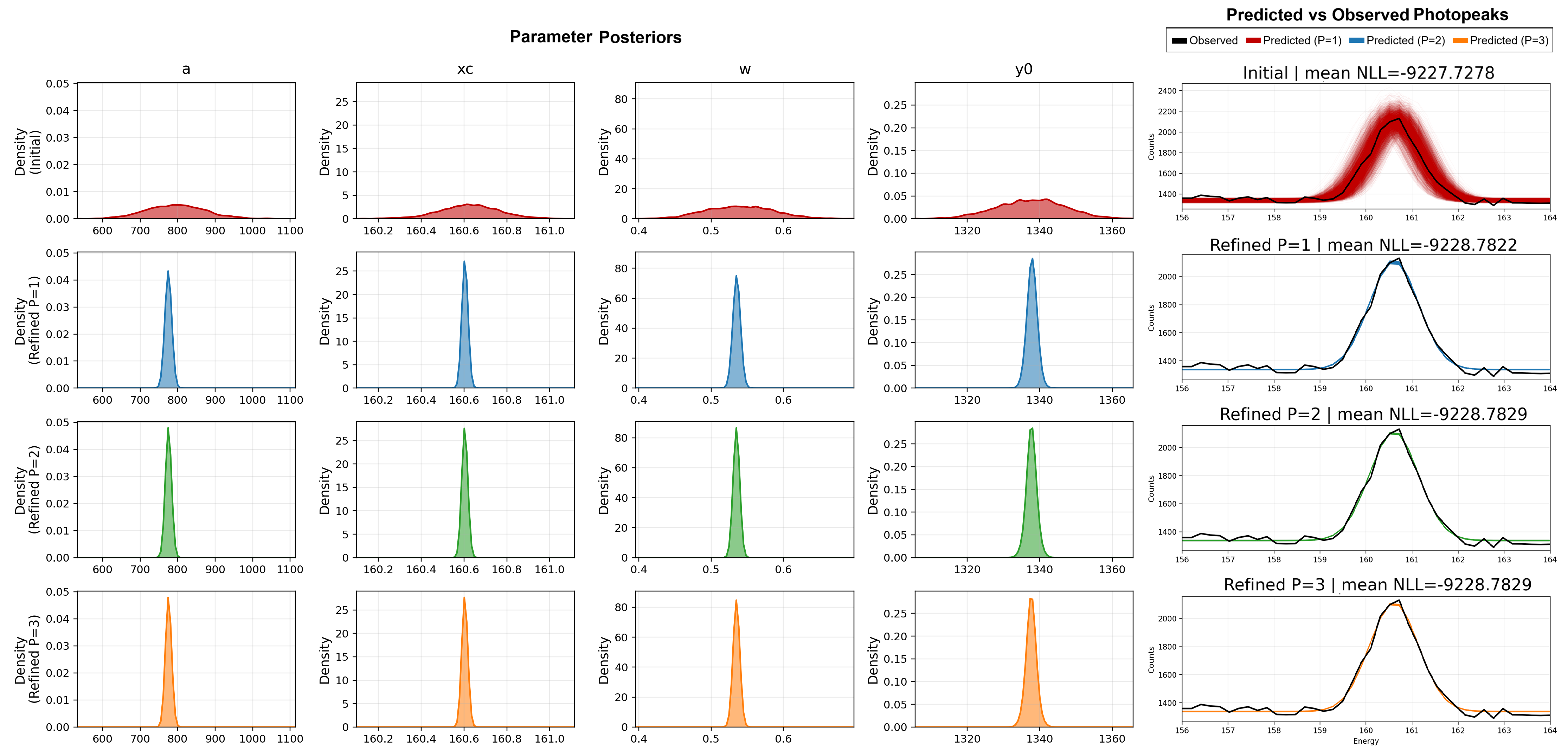}}
    \caption{
     Predictions on gamma-ray spectroscopy data using the refinement framework in this work. The first 4 columns in each row show predicted posteriors of parameters describing a photopeak for a sample Ba-133. The last column shows the predicted photopeak profiles corresponding to the predicted parameters compared to the observed photopeak. The NLL improves as number of parameters are increased in the refinement and converges at the 3-parameter stage ($P$ =3).
    }
    \label{fig:gamma_retrieval}
  \end{center}
\end{figure}

\begin{figure}[ht]
  \vskip 0.2in
  \begin{center}
    \centerline{\includegraphics[width=\columnwidth]{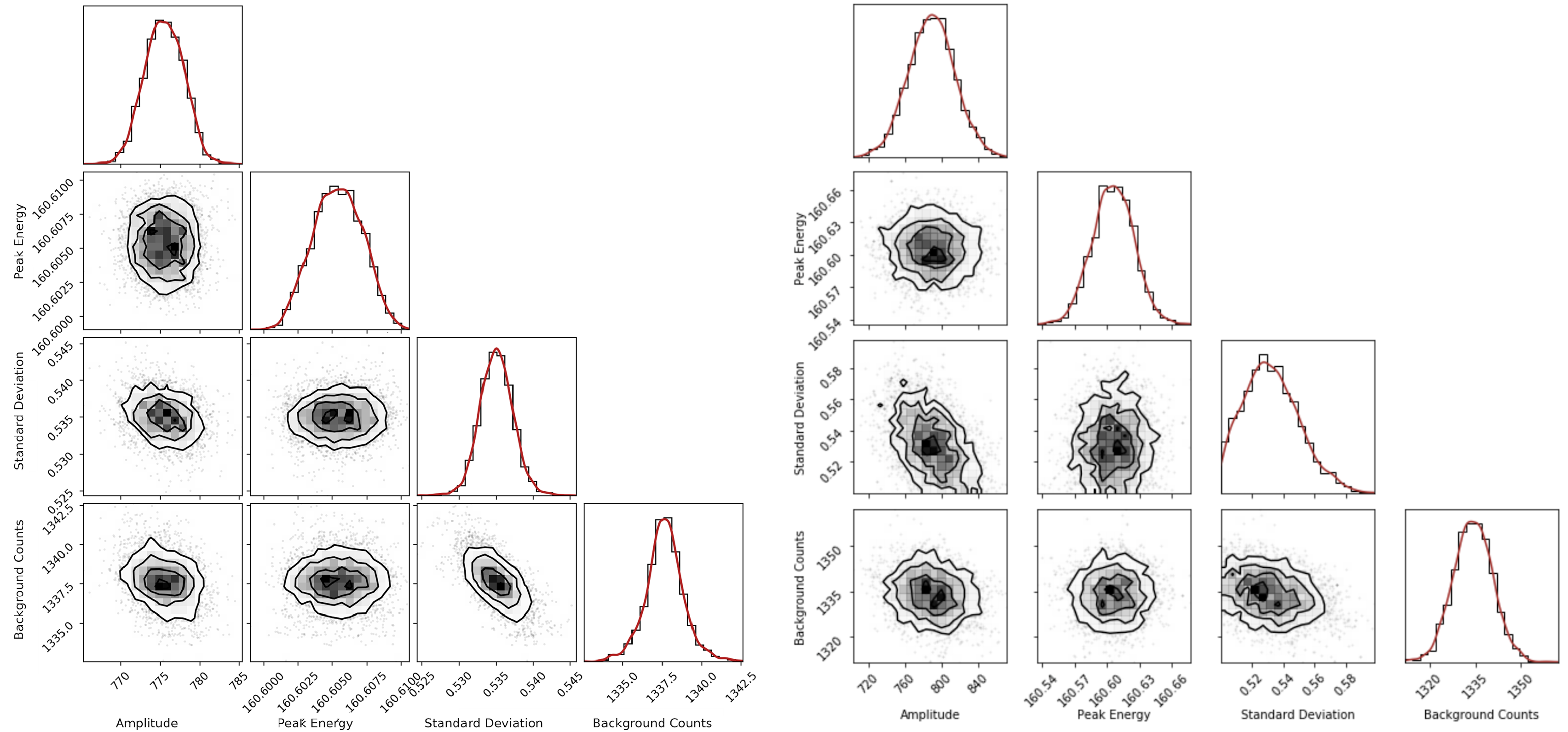}}
    \caption{
Joint and marginal posterior distributions for the Ba-133 photopeak
parameters.
\textbf{Left:} Posterior obtained from embedding-based retrieval followed by
local refinement.
\textbf{Right:} Reference posterior from MCMC sampling \cite{moghani2025sampling}.
Both posteriors exhibit the same dominant correlations, including the
amplitude–width degeneracy characteristic of Gaussian photopeak models.
}
    
    \label{fig:gamma_posteriors}
  \end{center}
\end{figure}

Figure~\ref{fig:gamma_posteriors} shows posterior marginals and predicted
photopeaks for the Ba-133 inference task. The initial retrieval ensemble
(top row) produces broad but correctly centered parameter distributions for
$(a, x_c, w, y_0)$. Local refinement sharpens improves data
agreement (see Figure ~\ref{fig:gamma_retrieval}).

Most posterior contraction occurs at first parameter, $P=1$, with minimal change for $P=2$ and $P=3$ by jointly perturbing 2 and 3 parameters respectively. Correspondingly, the mean Poisson NLL improves from $-9227.73$ (initial) to $-9228.78$ after refinement, with differences across $P$ below $10^{-2}$. Now, predicted spectra after the refinement closely match the observed photopeak. 

Figure~\ref{fig:gamma_posteriors} compares the refined posterior to the MCMC reference. The refined posterior recovers the same high-probability region in parameter space and
exhibits close agreement in the one-dimensional marginals for all parameters.
While differences remain in the joint structure, the refined posterior
captures the correct parameter scales and central values identified by MCMC.

\end{document}